\documentclass{article}

\PassOptionsToPackage{numbers, compress}{natbib}

    \usepackage[preprint]{neurips_2020}

\usepackage{appendix}
\usepackage[utf8]{inputenc} 
\usepackage[T1]{fontenc}    
\usepackage{hyperref}       
\usepackage{url}            
\usepackage{booktabs}       
\usepackage{amsfonts}       
\usepackage{amsmath}
\usepackage{nicefrac}       
\usepackage{microtype}      
\usepackage[pdftex]{graphicx}
\usepackage{algorithm}
\usepackage{algorithmic}
\usepackage{subcaption}
\usepackage{bbm}
\usepackage[dvipsnames]{xcolor}
\usepackage{cleveref}
\usepackage{enumitem}
\usepackage{wrapfig}

\newcommand{\G}[1]{G(#1)} 
\newcommand{\D}[1]{D(#1)} 
\newcommand{\Dh}[1]{D_h(#1)} 
\newcommand{\proj}[1]{f(#1)} 
\newcommand{\T}{\mathcal{T}} 
\newcommand{\bcr}{\text{BCR}} 
\newcommand{\cntr}{\text{Cntr}} 
\newcommand{\img}{\mathcal{I}} 
\newcommand{\real}{\text{real}} 
\newcommand{\fake}{\text{fake}} 

\newcommand{\augstr}{\lambda_\text{aug}}

\newcommand\myh{3.2cm}
\newcommand\myhh{3.3cm}

\title{Image Augmentations for GAN Training}

\author{%
  Zhengli Zhao\thanks{Work done as an intern on Google Brain team.} \\
  UC Irvine\\
  \texttt{zhengliz@uci.edu} \\
\And
  Zizhao Zhang, Ting Chen \\
  Google Research\\
\And
  Sameer Singh \\
  UC Irvine \\
\And
  Han Zhang \\
  Google Research \\
}

\begin{document}

\maketitle

\begin{abstract}
Data augmentations have been widely studied to improve the accuracy and robustness of classifiers. However, the potential of image augmentation in improving GAN models for image synthesis has not been thoroughly investigated in previous studies.  In this work, we systematically study the effectiveness of various existing augmentation techniques for GAN training in a variety of settings. We provide insights and guidelines on how to augment images for both vanilla GANs and GANs with regularizations, improving the fidelity of the generated images substantially. 
Surprisingly, we find that vanilla GANs attain generation quality on par with recent state-of-the-art results if we use augmentations on \emph{both} real and generated images.
When this GAN training is combined with other augmentation-based regularization techniques, such as contrastive loss and consistency regularization, the augmentations further improve the quality of generated images.
We provide new state-of-the-art results for conditional generation on CIFAR-10 with both consistency loss and contrastive loss as additional regularizations.

\end{abstract}

\section{Introduction}\label{sec:intro}

Data Augmentation has played an important role in deep representation learning. It increases the amount of training data in a way that is natural/useful for the domain, and thus reduces over-fitting when training deep neural networks with millions of parameters. In the image domain, a variety of augmentation techniques have been proposed to improve the performance of different visual recognition tasks such as image classification~\citep{AlexNet, ResNet, SIMCLR}, object detection~\citep{FasterRCNN, AugforDetect}, and semantic segmentation~\citep{DeepLab,MaskRCNN}. The augmentation strategies also range from the basic operations like random crop and horizontal flip to more sophisticated handcrafted operations~\citep{CUTOUT, CUTMIX,MIXUP,AugMix}, or even the strategies directly learned by the neural network~\citep{AutoAug, ADVAutoAug}. However, previous studies have not provided a systematic study of the impact of the data augmentation strategies for deep generative models, especially for image generation using Generative Adversarial Networks (GANs)~\citep{GAN}, making it unclear how to select the augmentation techniques, which images to apply them to, how to incorporate them in the loss, and therefore, how useful they actually are.

Compared with visual recognition tasks, making the right choices for the augmentation strategies for image generation is substantially more challenging. 
Since most of the GAN models only augment real images as they are fed into the discriminator, the discriminator mistakenly learns that the augmented images are part of the image distribution. 
The generator thus learns to produce images with undesired augmentation artifacts, such as cutout regions and jittered color if advanced image augmentation operations are used ~\citep{CRGAN, ICRGAN}. 
Therefore, the state-of-the-art GAN models~\citep{DCGAN, StackGAN, SAGAN, BIGGAN, StyleGAN} prefer to use random crop and flip as the only augmentation strategies. 
In unsupervised and self-supervised learning communities, image augmentation becomes a critical component of consistency regularization~\citep{laine2016temporal, CONSISTENCY, UDACONSISTENCY}. 
Recently, \citet{CRGAN} studied the effect of several augmentation strategies when applying consistency regularization in GANs, where they enforced the discriminator outputs to be unchanged when applying several perturbations to the real images.
\citet{ICRGAN} have further improved the generation quality by adding augmentations on both the generated samples and real images. 
However, it remains unclear about the best strategy to use augmented data in GANs: Which image augmentation operation is more effective in GANs? Is it necessary to add augmentations in generated images as in ~\citet{ICRGAN}? 
Should we always couple augmentation with consistency loss like by \citet{CRGAN}?
Can we apply augmentations together with other loss constraints besides consistency?

In this paper, we comprehensively evaluate a broad set of common image transformations as augmentations in GANs.
We first apply them in the conventional way---only to the real images fed into the discriminator. We vary the strength for each augmentation and compare the generated samples in FID~\citep{FID} to demonstrate the efficacy and robustness for each augmentation. 
We then evaluation the quality of generation when we add each augmentation to both real images \emph{and} samples generated during GAN training. 
Through extensive experiments, we conclude that only augmenting real images is ineffective for GAN training, whereas augmenting both real and generated images consistently improve GAN generation performance significantly.
We further improve the results by adding consistency regularization~\citep{CRGAN, ICRGAN} on top of augmentation strategies and demonstrate such regularization is necessary to achieve superior results. 
Finally, we apply consistency loss together with contrastive loss, and show that combining regularization constraints with the best augmentation strategy achieves the new state-of-the-art results. 

In summary, our contributions are as follows:

\begin{itemize}

\item We conduct extensive experiments to assess the efficacy and robustness for different augmentations in GANs to guide researchers and practitioners for future exploration.

\item We provide a thorough empirical analysis to demonstrate augmentations should be added to both real and fake images, with the help of which we improve the FID of vanilla BigGAN to 11.03, outperforming BigGAN with consistency regularization in ~\citet{CRGAN}.

\item We demonstrate that adding regularization on top of augmentation furthers boost the quality. Consistency loss compares favorably against contrastive loss as the regularization approach.

\item We achieve new state-of-the-art for image generation by applying contrastive loss and consistency loss on top of the best augmentation we find. We improve the  state-of-the-art FID of conditional image generation for CIFAR-10 from 9.21 to 8.30. 

\end{itemize}

\section{Augmentations and Experiment Settings} \label{sec:aug}

\begin{figure}[tb]
\begin{subfigure}{0.162\linewidth}
\frame{\includegraphics[width=\linewidth]{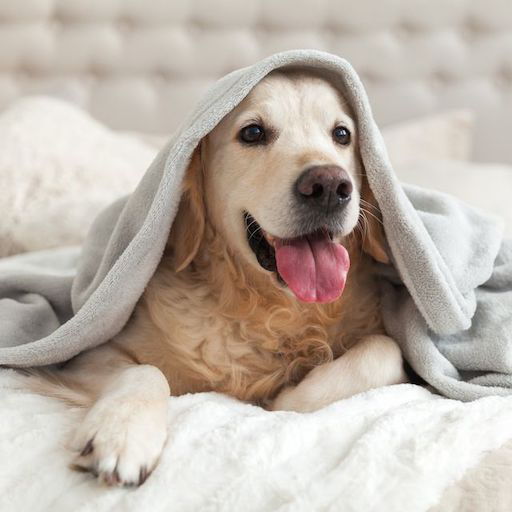}}
\caption*{Original Image}\end{subfigure}  
\begin{subfigure}{0.162\linewidth}
\frame{\includegraphics[width=\linewidth]{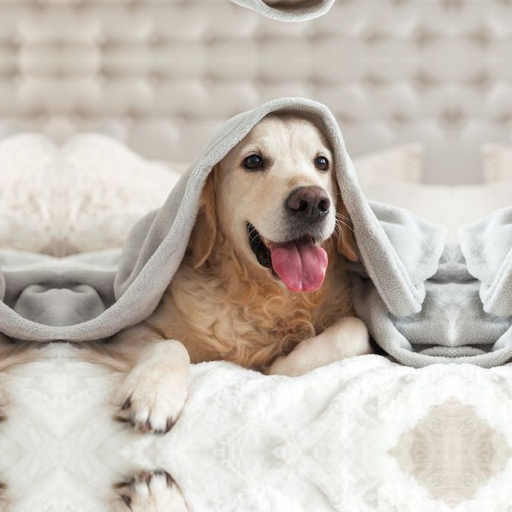}}
\caption*{ZoomOut}\end{subfigure}  
\begin{subfigure}{0.162\linewidth}
\frame{\includegraphics[width=\linewidth]{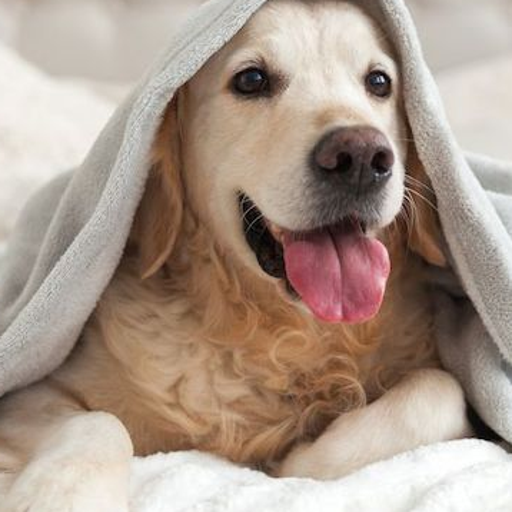}}
\caption*{ZoomIn}\end{subfigure}  
\begin{subfigure}{0.162\linewidth}
\frame{\includegraphics[width=\linewidth]{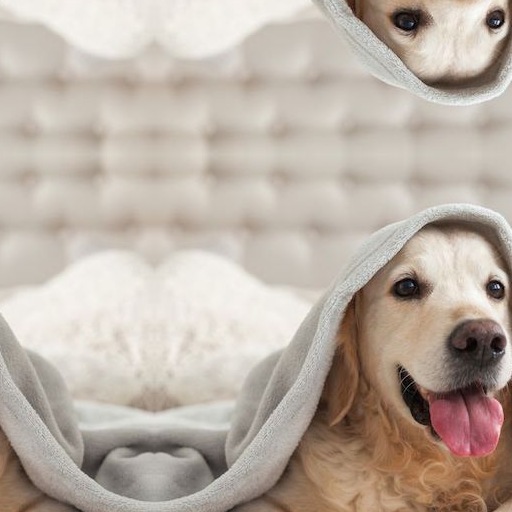}}
\caption*{Translation}\end{subfigure}  
\begin{subfigure}{0.162\linewidth}
\frame{\includegraphics[width=\linewidth]{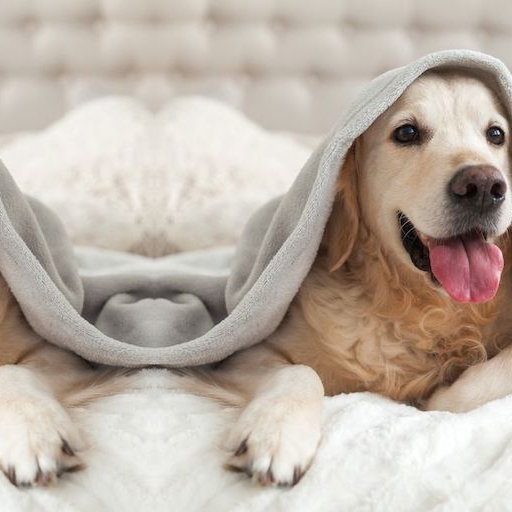}}
\caption*{TranslationX}\end{subfigure}  
\begin{subfigure}{0.162\linewidth}
\frame{\includegraphics[width=\linewidth]{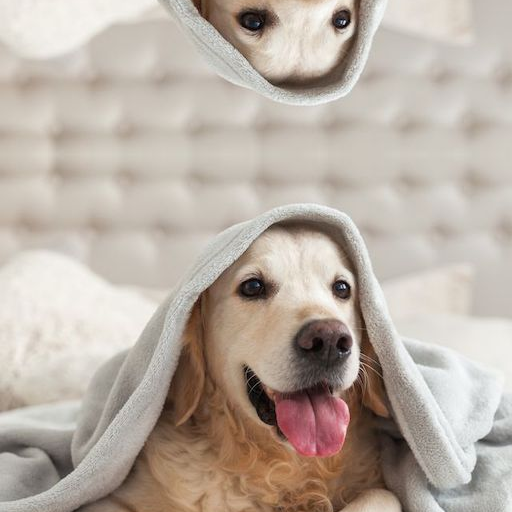}}
\caption*{TranslationY}\end{subfigure}  
\begin{subfigure}{0.162\linewidth}
\frame{\includegraphics[width=\linewidth]{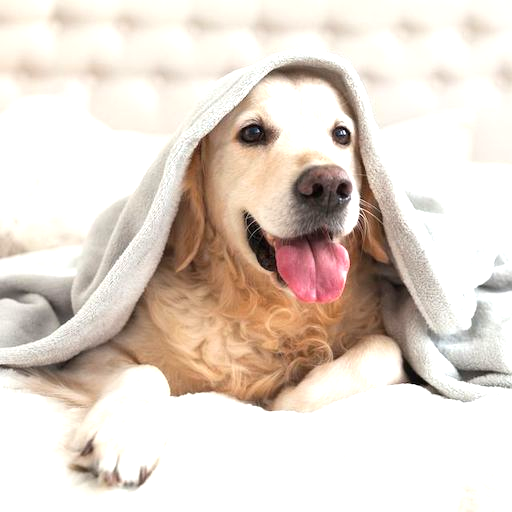}}
\caption*{Brightness}\end{subfigure}  
\begin{subfigure}{0.162\linewidth}
\frame{\includegraphics[width=\linewidth]{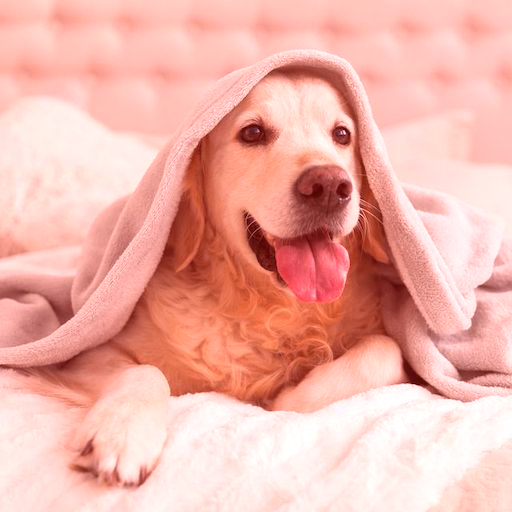}}
\caption*{Redness}\end{subfigure}  
\begin{subfigure}{0.162\linewidth}
\frame{\includegraphics[width=\linewidth]{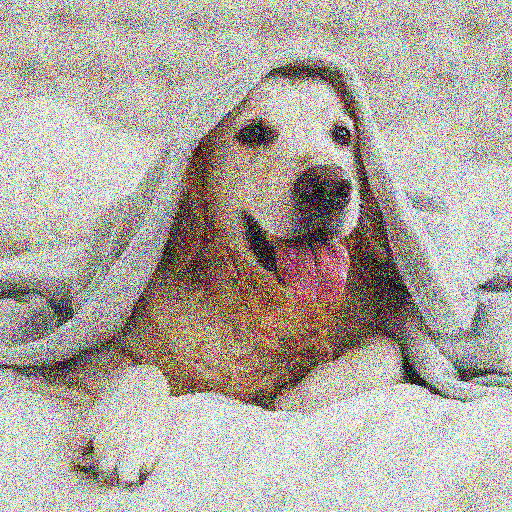}}
\caption*{InstanceNoise}\end{subfigure}  
\begin{subfigure}{0.162\linewidth}
\frame{\includegraphics[width=\linewidth]{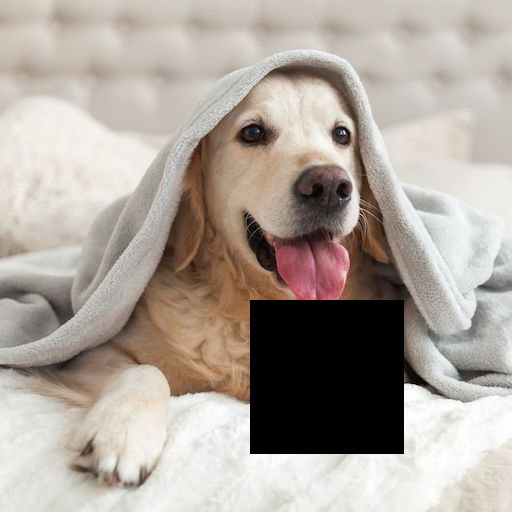}}
\caption*{CutOut\citep{CUTOUT}}\end{subfigure}  
\begin{subfigure}{0.162\linewidth}
\frame{\includegraphics[width=\linewidth]{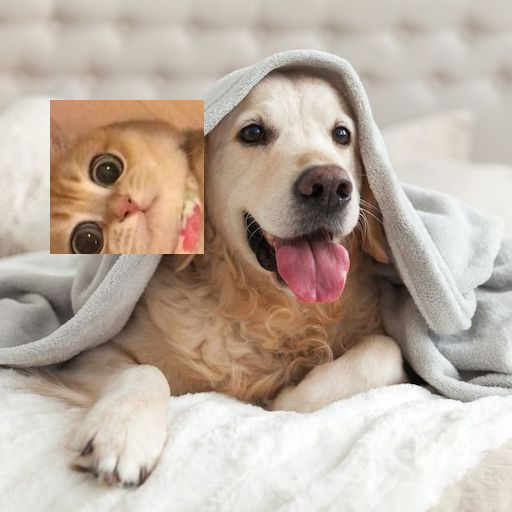}}
\caption*{CutMix\citep{CUTMIX}}\end{subfigure}  
\begin{subfigure}{0.162\linewidth}
\frame{\includegraphics[width=\linewidth]{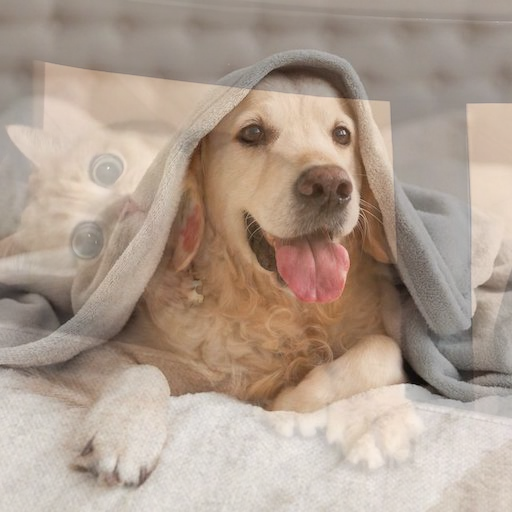}}
\caption*{MixUp\citep{MIXUP}}\end{subfigure}  
\caption{Different augmentation techniques applied to the original image.
}
\label{fig:aug}
\end{figure}

We first introduce the image augmentation techniques we study in this paper,
and then elaborate on the datasets, GAN architectures, hyperparameters, and evaluation metric used in the experiments.

\textbf{Image Augmentations}.
Our goal is to investigate how each image operation performs in the GAN setting. Therefore, instead of chaining augmentations~\citep{AutoAug, RandAug}, we have selected 10 basic image augmentation operations and 3 advanced image augmentation techniques as the candidates $\T$, which are illustrated in Figure~\ref{fig:aug}.
The original image $\img_0$ of size $(H, W)$ is normalized with the pixel range in $[0, 1]$.  
For each augmentation $t \sim \T$, the strength $\augstr$ is chosen uniformly in the space ranging from the weakest to the strongest one. We note that $t(\img_0)$ is the augmented image and 
we detail each augmentation in Section~\ref{sec:appx:aug} in the appendix.

\textbf{Data}.
We validate all the augmentation strategies on the CIFAR-10 dataset~\citep{CIFAR}, 
which consists of 60K of 32x32 images in 10 classes. 
The size of this dataset is suitable for a large scale study in GANs~\citep{LucicKMGB18,compare_gan}. 
Following previous work, we use 50K images for training and 10K for evaluation.

\textbf{Evaluation metric}.
We adopt Fr\'echet Inception Distance (FID)~\citep{FID} as the metric for quantitative evaluation. We admit that better (i.e., lower) FID does not always imply better image quality, but FID is proved to be more consistent with human evaluation and widely used for GAN evaluation. 
Following~\citet{compare_gan}, we carry out experiments with different random seeds, and aggregate all runs and report FID of the top 15\% trained models. FID is calculated on the test dataset with 10K generated samples and 10K test images.

\textbf{GAN architectures and training hyperparameters}.
The search space for GANs is prohibitively large. As our main purpose is to evaluate different augmentation strategies, we select two commonly used settings and GANs architectures for evaluation, namely SNDCGAN~\citep{Miyato18a} for unconditional image generation and BigGAN~\citep{BIGGAN} for conditional image generation. As in previous work~\citep{compare_gan, CRGAN}, we train SNDCGAN with batch size 64 and the total training step is 200k. For conditional BigGAN, we set batch size as 256 and train for 100k steps. We choose hinge loss~\citep{lim2017, Tran2017} for all the experiments. More details of hyperparameter settings can be found in appendix.

We first study augmentations on \emph{vanilla} SNDCGAN and BigGAN without additional regularizations in Section~\ref{sec:vanilla}, then move onto these GANs with additional regularizations that utilize augmentations, namely
\emph{consistency regularization} (detailed in Section~\ref{sec:bcr}) and \emph{contrastive loss} (detailed in Section~\ref{sec:cntr}).

\section{Effect of Image Augmentations for Vanilla GAN}
\label{sec:vanilla}

In this section, we first study the effect of image augmentations when used conventionally---only augmenting real images. Then we propose and study a novel way 
where both real and generated images are augmented before fed into the discriminator,
which substantially improves GANs' performance.

\subsection{Augmenting Only Real Images Does Not Help with GAN Training} \label{sec:vanilla_r}

\begin{figure}[hb] 
  \centering
    \includegraphics[width=1\linewidth]{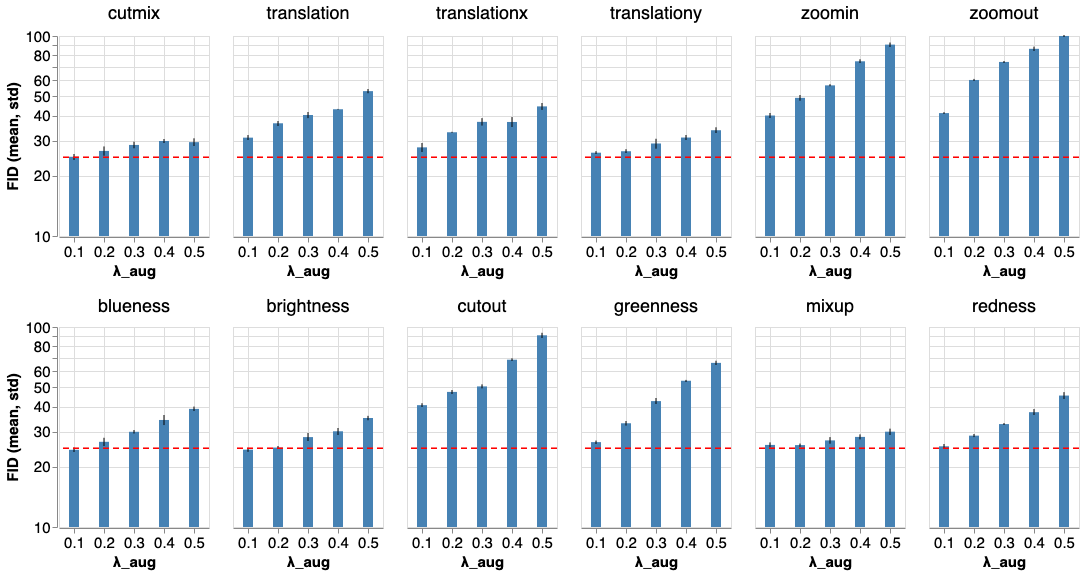}
  \caption{FID comparisons of SNDCGAN trained on augmented real images only. 
  It shows only augmenting real images is not helpful with vanilla GAN training, 
  which is consistent with the result in Section 4.1 of~\citet{CRGAN}.
  Corresponding plots of BigGAN results are in the appendix.
  }
\label{fig:aug_only_real_vanilla}
\end{figure}

We first compare the effect of image augmentations when only applied to the real images, the de-facto way of image augmentations in GANs~\citep{DCGAN, BIGGAN, styleganv2}. Figure~\ref{fig:aug_only_real_vanilla} illustrates the FID of the generated images with different strengths of each augmentation. We find \emph{augmenting only real images in GANs worsens the FID regardless of the augmentation strengths or strategies}. 
For example, the baseline SNDCGAN trained without any image augmentation achieves 24.73 in FID~\citep{CRGAN},  while \emph{translation}, even with its smallest strength, gets 31.03. 
Moreover, FID increases monotonically as we increase the strength of the augmentation. 
This conclusion is surprising given the wide adoption of this conventional image augmentations in GANs. We note that the discriminator is likely to view the augmented data as part of the data distribution in such case.
As shown by \Cref{fig:artifacts:brightness,fig:artifacts:blueness,fig:artifacts:zoomin,fig:artifacts:cutout} in the appendix, the generated images are prone to contain augmentation artifacts. 
Since FID is calculating the feature distance between generated samples and unaugmented real images, we believe the augmented artifacts in the synthesized samples are the underlying reason for the inferior FID.

\subsection{Augmenting Both Real and Fake Images Improves GANs Consistently} \label{sec:vanilla_rf}

\begin{figure}[tb]
\begin{subfigure}{0.16\linewidth}
\includegraphics[height=\myh]{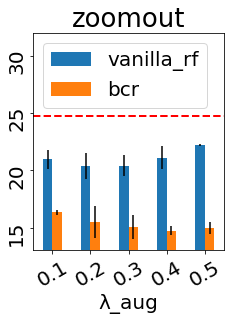}\end{subfigure}
\begin{subfigure}{0.16\linewidth}
\includegraphics[height=\myh]{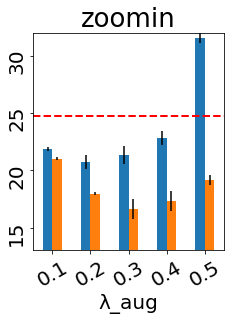}\end{subfigure}
\begin{subfigure}{0.16\linewidth}
\includegraphics[height=\myh]{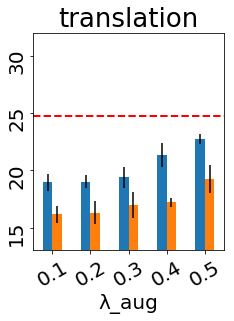}\end{subfigure}
\begin{subfigure}{0.16\linewidth}
\includegraphics[height=\myh]{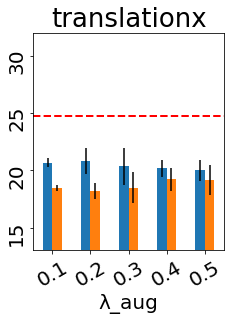}\end{subfigure}
\begin{subfigure}{0.16\linewidth}
\includegraphics[height=\myh]{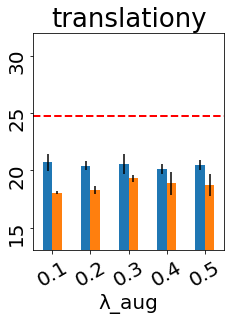}\end{subfigure}
\begin{subfigure}{0.16\linewidth}
\includegraphics[height=\myh]{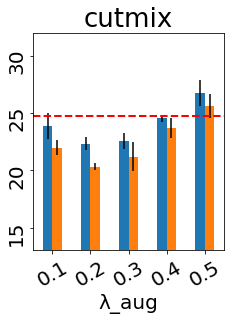}\end{subfigure}
\begin{subfigure}{0.16\linewidth}
\includegraphics[height=\myhh]{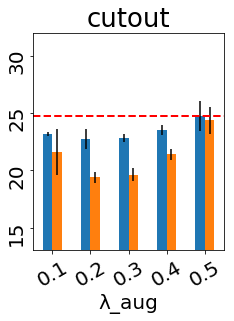}\end{subfigure}
\begin{subfigure}{0.16\linewidth}
\includegraphics[height=\myhh]{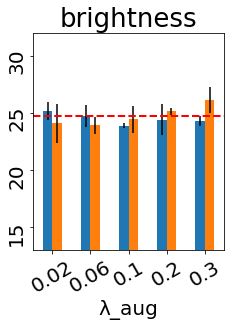}\end{subfigure}
\begin{subfigure}{0.16\linewidth}
\includegraphics[height=\myhh]{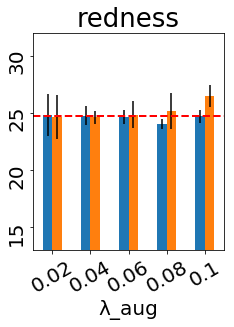}\end{subfigure}
\begin{subfigure}{0.16\linewidth}
\includegraphics[height=\myhh]{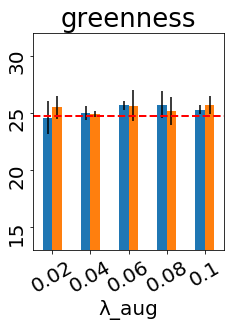}\end{subfigure}
\begin{subfigure}{0.16\linewidth}
\includegraphics[height=\myhh]{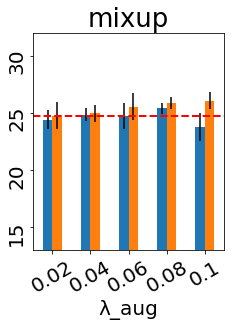}\end{subfigure}
\begin{subfigure}{0.16\linewidth}
\includegraphics[height=\myhh]{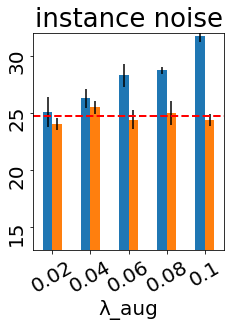}\end{subfigure}
\caption{FID comparisons of SNDCGAN on CIFAR-10.
The red dashed horizontal line shows the baseline FID=24.73 of SNDCGAN trained without data augmentation.
`vanilla\_rf' (Section~\ref{sec:vanilla_rf}) represents training vanilla SNDCGAN and augmenting both real images as well as generated fake images concurrently before fed into the discriminator.
And `bcr' (Section~\ref{sec:bcr}) corresponds to training SNDCGAN with Balanced Consistency Regularization on augmented real and fake images.
This figure can be utilized as general guidelines for training GAN with augmentations.
The main implications are:
(1) Simply augmenting real and fake images can make the vanilla GAN's performance on par with recent proposed CR-GAN~\citep{CRGAN}.
(2) With the help of BCR on augmented real and fake images, the generation fidelity can be improved by even larger margins.
(3) Spatial augmentations outperform visual augmentations.
(4) Augmentations that result in images out of the natural data manifold, e.g. InstanceNoise, cannot help with improving GAN performance.
}
\label{fig:sndcgan_vanilla_bcr}
\end{figure}

Based on the above observation, it is natural to wonder whether augmenting generated images in the same way before feeding them into the discriminator can alleviate the problem. In this way, augmentation artifacts cannot be used to distinguish real and fake images by the discriminator. 

To evaluate the augmentation of synthetic images, we train SNDCGAN and BigGAN by augmenting both real images as well as generated images concurrently before feeding them into the discriminator during training.
Different from augmenting real images, we keep the gradients for augmented generated images to train the generator.
The discriminator is now trained to differentiate between the augmented real image $t(\img_\real)$ and the augmented fake image $t(\G{z})$.
We present the generation FID of SNDCGAN and BigGAN in \Cref{fig:sndcgan_vanilla_bcr,fig:biggan_vanilla_bcr} (denoted as `vanilla\_rf'), where the horizontal lines show the baseline FIDs without any augmentations. 
As illustrated by Figure~\ref{fig:sndcgan_vanilla_bcr}, this new augmentation strategy considerably improves the FID for different augmentations with varying strengths. By comparing the results in Figure~\ref{fig:sndcgan_vanilla_bcr} and Figure~\ref{fig:aug_only_real_vanilla},
we conclude that \emph{augmenting both real and fake images can substantially improves the generation performance of GAN}. 
Moreover, for SNDCGAN, we find the best FID 18.94 achieved by \emph{translation} of strength 0.1 is comparable to the FID 18.72 reported in \citet{CRGAN} with consistency regularization only on augmented real images. This observation holds for BigGAN as well, where we get FID 11.03 and the FID of CRGAN~\citep{CRGAN} is 11.48.
These results suggest that image augmentations for both real and fake images considerably improve the training of vanilla GANs,
which has not been studied by previous work, to our best knowledge.

We compare the effectiveness of augmentation operations in \Cref{fig:sndcgan_vanilla_bcr,fig:biggan_vanilla_bcr}.
The operations in the top row such as \emph{translation}, \emph{zoomin}, and \emph{zoomout}, are much more effective than the operations in the bottom rows, such as \emph{brightness}, \emph{colorness}, and \emph{mixup}. We conclude that augmentations that result in \emph{spatial} changes improve the GAN performance more than those that induce mostly \emph{visual} changes.


\subsection{Augmentations Increase the Support Overlap between Real and Fake Distributions}

In this section, we investigate the reasons why augmenting both real and fake images improves GAN performance considerably.
Roughly, GANs' objective corresponds to making the generated image distribution close to real image distribution.
However, as mentioned by previous work~\citep{NOISE, ArjovskyB17}, the difficulty of training GANs stems from these two being concentrated distributions whose support do not overlap:
the real image distribution is often assumed to concentrate on or around a low-dimensional manifold, and similarly, generated image distribution is degenerate by construction. Therefore, \citet{NOISE} propose to add \emph{instance noise} (i.e., Gaussian Noise) as augmentation for both real images and fakes image to increase the overlap of support between these two distributions. We argue that other semantic-preserving image augmentations have a similar effect to increase the overlap, and are much more effective for image generation.

\begin{figure}[tb]
  \centering
    \includegraphics[trim={0 0 0 0.4cm},clip,width=1\linewidth]{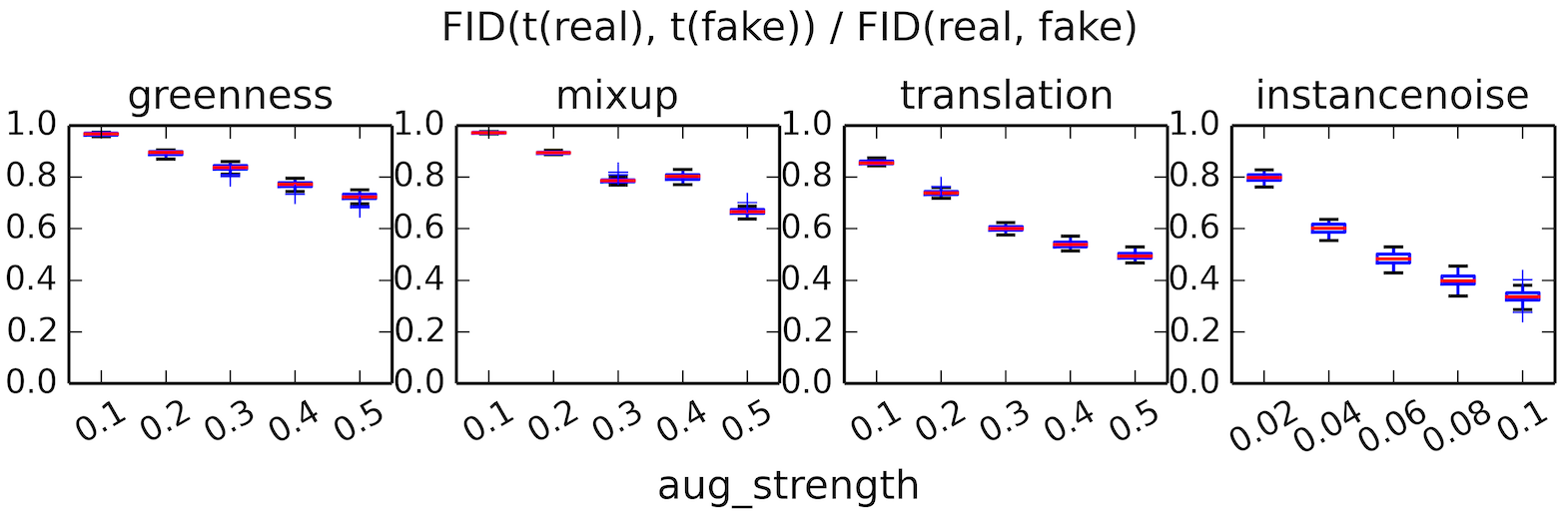}
  \caption{Distances between real and fake distributions with different augmentations.
  Note that we report $\text{FID}(\img_\real, \img_\fake)$ normally throughout the paper.
  While here to show the changes of real and fake image distributions with augmentations, we also calculate $\text{FID}(t(\img_\real), t(\img_\fake))$ and present its fraction over normal FID as y-axis.
  The Frechet Inception Distance between real and fake images gets smaller with augmentations, while stronger augmentations result in more distribution overlaps.
  }
 \label{fig:aug_decrease_fid}
\end{figure}

In Figure~\ref{fig:aug_decrease_fid}, we show that augmentations $t \sim \T$ can lower FID between augmented $t(\img_\real)$ and 
$t(\img_\fake)$, which indicates that the support of image distribution and the support of model distribution have more overlaps with augmentations.
However, not all augmentations or strengths can improve the quality of generated images, which suggests naively pulling distribution together may not always improve the generation quality. 
We hypothesize certain types of augmentations and augmentations of high strengths can result in images that are far away from the natural image distribution; we leave the theoretical justification for future work.

\section{Effect of Image Augmentations for Consistency Regularized GANs} \label{sec:bcr}

\begin{figure}[tb]
\begin{subfigure}{0.16\linewidth}
\includegraphics[height=\myh]{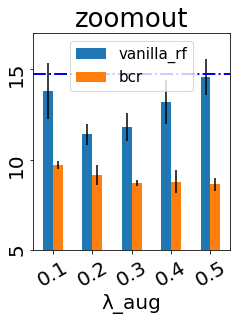}\end{subfigure}
\begin{subfigure}{0.16\linewidth}
\includegraphics[height=\myh]{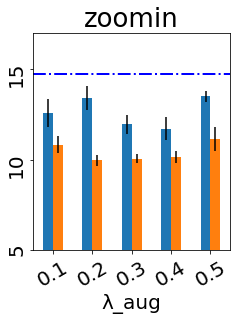}\end{subfigure}
\begin{subfigure}{0.16\linewidth}
\includegraphics[height=\myh]{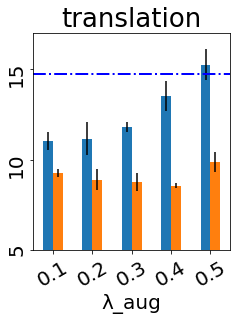}\end{subfigure}
\begin{subfigure}{0.16\linewidth}
\includegraphics[height=\myh]{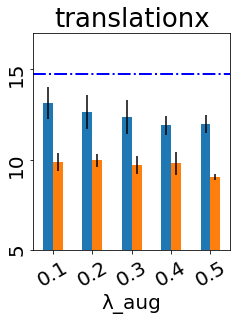}\end{subfigure}
\begin{subfigure}{0.16\linewidth}
\includegraphics[height=\myh]{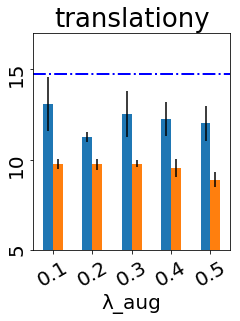}\end{subfigure}
\begin{subfigure}{0.16\linewidth}
\includegraphics[height=\myh]{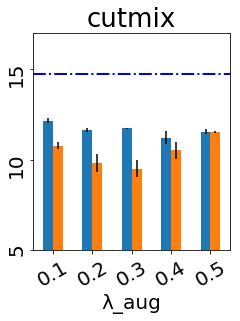}\end{subfigure}
\begin{subfigure}{0.16\linewidth}
\includegraphics[height=\myh]{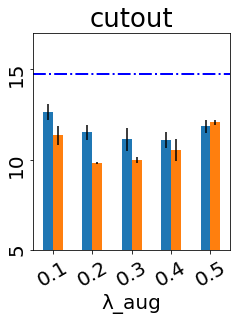}\end{subfigure}
\begin{subfigure}{0.16\linewidth}
\includegraphics[height=\myh]{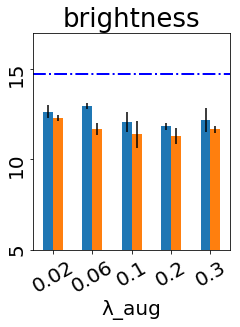}\end{subfigure}
\begin{subfigure}{0.16\linewidth}
\includegraphics[height=\myh]{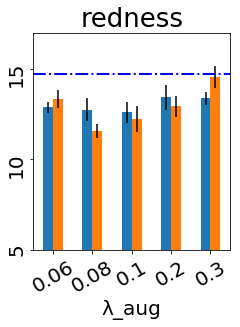}\end{subfigure}
\begin{subfigure}{0.16\linewidth}
\includegraphics[height=\myh]{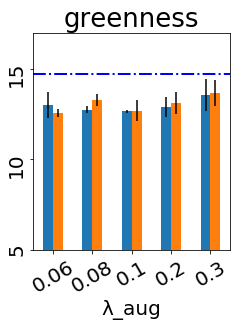}\end{subfigure}
\begin{subfigure}{0.16\linewidth}
\includegraphics[height=\myh]{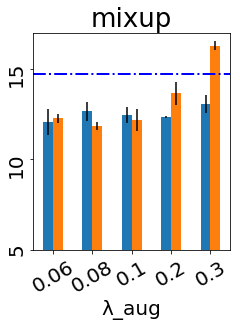}\end{subfigure}
\begin{subfigure}{0.16\linewidth}
\includegraphics[height=\myh]{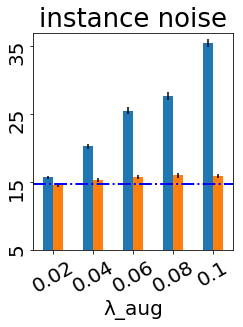}\end{subfigure}
\caption{FID mean and std of BigGAN on CIFAR-10.
The blue dashed horizontal line shows the baseline FID=14.73 of BigGAN trained without augmentation.
`vanilla\_rf' (Section~\ref{sec:vanilla_rf}) represents training vanilla BigGAN with both real and fake images augmented.
`bcr' (Section~\ref{sec:bcr}) corresponds to training BigGAN with BCR on augmented real and fake images.
This figure can be utilized as general guidelines for training GAN with augmentations, sharing similar  implications as in Figure~\ref{fig:sndcgan_vanilla_bcr}.
}
\label{fig:biggan_vanilla_bcr}
\end{figure}

We now turn to more advanced regularized GANs that built on their usage of augmentations. 
Consistency Regularized GAN (CR-GAN) \citep{CRGAN} has demonstrated that consistency regularization can significantly improve GAN training stability and generation performance.
\citet{ICRGAN} improves this method by introducing Balanced Consistency Regularization (BCR), which applying BCR to both real and fake images. Both methods requires images to be augmented for processing, and we briefly summarize BCR-GAN with Algorithm~\ref{alg:bcr} in the appendix. 

However, neither of the works studies the impact and importance of individual augmentation and only very basic geometric transformations are used as augmentation. We believe an in-depth analysis of augmentation techniques can strengthen the down-stream applications of consistency regularization in GANs. 
Here we mainly focus on analyzing the efficacy of different augmentations on BCR-GAN.
We set the BCR strength  $\lambda_\bcr = 10$ in Algorithm~\ref{alg:bcr} according to the best practice.
We present the generation FID of SNDCGAN and BigGAN with BCR on augmented real and fake images in \Cref{fig:sndcgan_vanilla_bcr,fig:biggan_vanilla_bcr} (denoted as `bcr'), where the horizontal lines show the baseline FIDs without any augmentation.
Experimental results suggest that \emph{consistency regularization on augmentations for real and fake images can further boost the generation performance.} 

More importantly, we can also significantly outperform the state of the art by carefully selecting the augmentation type and strength.
For SNDCGAN, the best FID 14.72 is with \emph{zoomout} of strength 0.4, while the corresponding FID reported in~\citet{ICRGAN} is 15.87 where basic translation of 4 pixels and flipping are applied.
The best BigGAN FID 8.65 is with \emph{translation} of strength 0.4, outperforming the corresponding FID 9.21 reported in~\citet{ICRGAN}.

Similarly as in Section~\ref{sec:vanilla_rf}, augmentation techniques can be roughly categorized into two groups, in the descending order of effectiveness: 
\textbf{spatial} transforms, \emph{zoomout}, \emph{zoomin}, \emph{translation}, \emph{translationx}, \emph{translationy}, \emph{cutout}, \emph{cutmix};
and \textbf{visual} transforms, \emph{brightness}, \emph{redness}, \emph{greenness}, \emph{blueness}, \emph{mixup}.
Spatial transforms, which retain the major content while introducing spatial variances,
can substantially improve GAN performance together with BCR.
On the other hand, \emph{instance noise}~\citep{NOISE}, which may be able to help stabilize GAN training, cannot improve generation performance.


\section{Effect of Images Augmentations for GANs with Contrastive Loss}
\label{sec:cntr}

Image augmentation is also an essential component of contrastive learning, which has recently led to substantially improved performance on self-supervised learning~\citep{SIMCLR,he2019momentum}. Given the success of contrastive loss for representation learning and the success of consistency regularization in GANs, it naturally raises the question of whether adding such a regularization term helps in training GANs?
In this section, we first demonstrate how we apply contrastive loss (CntrLoss) to regularizing GAN training.
Then we analyze on how the performance of Cntr-GAN is affected by different augmentations,
including variations of an augmentation set in existing work~\citep{SIMCLR}.

\paragraph{Contrastive Loss for GAN Training} 

The contrastive loss was originally introduced by \citet{hadsell2006dimensionality} in such a way that 
corresponding positive pairs are pulled together while negative pairs are pushed apart.
Here we propose \textbf{Cntr-GAN},
where contrastive loss is applied to regularizing the discriminator on two random augmented copies of both real and fake images. 
CntrLoss encourages the discriminator to push different image representations apart, while drawing augmentations of the same image closer. 
Due to space limit, we detail the CntrLoss in Appendix~\ref{sec:appx:cntr} and illustrate how our Cntr-GAN is trained with augmenting both real and fake images (Algorithm~\ref{alg:cntr}) in the appendix.

For augmentation techniques,
we adopt and sample the augmentation as described in \citet{SIMCLR}, referring it as \emph{simclr}. 
Details of \emph{simclr} augmentation can be found in the appendix (Section~\ref{sec:appx:aug}). 
Due to the preference of large batch size for CntrLoss, we mainly experiment on BigGAN which has higher model capacity. 
As shown in Table~\ref{tab:cntr_bcr}, Cntr-GAN outperforms baseline BigGAN without any augmentation, but is inferior to BCR-GAN.

\begin{wraptable}{r}{6cm}
\caption{BigGAN and regularizations.}
\centering
\small
\begin{tabular}{lcc}
\toprule
Regularization & FID & InceptionScore \\
\midrule
Vanilla     & 14.73         & 9.22~\citep{BIGGAN}   \\
\cntr       & 12.27         & 9.23                  \\
\bcr        & 9.21          & 9.29                  \\
\cntr+\bcr  & \textbf{8.30} & \textbf{9.41}         \\
\bottomrule
\end{tabular}
\label{tab:cntr_bcr}
\end{wraptable}

Since both BCR and CntrLoss utilize augmentations 
but are complementary in how they draw positive image pairs closer and push negative pairs apart,
we further experiment on regularizing BigGAN with both CntrLoss and BCR.
We are able to achieve new state-of-the-art $\text{FID}=8.30$ with $\lambda_\cntr=0.1, \lambda_\bcr=5$.
Table~\ref{tab:cntr_bcr} compares the performance of vanilla BigGAN against BigGAN with different regularizations on augmentations,
and Figure~\ref{fig:cntr_bcr} in the appendix shows how the strengths affect the results.
While BCR enforces the consistency loss directly on the discriminator logits,
with Cntr together, it further helps to learn better representations which can be reflected in generation performance eventually.

\paragraph{Cntr-GAN Benefits From Stronger Augmentations}

\begin{figure}[b]
  \centering
    \includegraphics[width=1\linewidth]{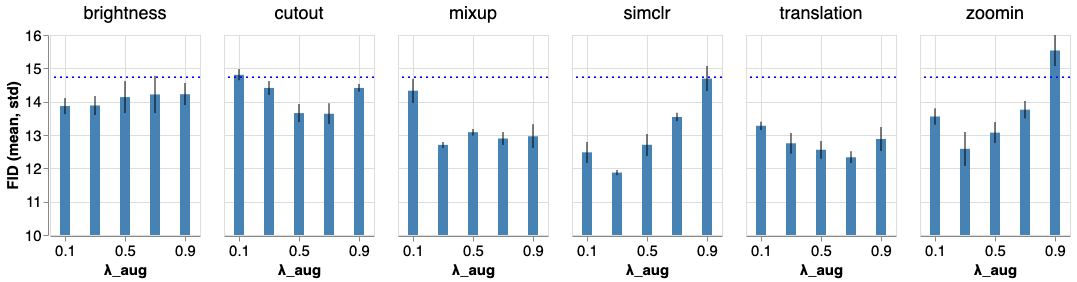}
  \caption{BigGAN regularized by CntrLoss with different image augmentations.
The blue dashed horizontal line shows the baseline FID=14.73 of BigGAN trained without augmentation. Here we adjust the strength of cropping-resizing in the default \emph{simclr}. Cntr-GAN consistently outperforms vanilla GAN with preferance on spatial augmentations.
}
\label{fig:aug_biggan_cntr}
\end{figure}

In Table~\ref{tab:cntr_bcr}, 
we adopt default augmenations in the literature for BCR~\citep{ICRGAN} and CntrLoss~\citep{SIMCLR}.
Now we further study which image transform used by \emph{simclr} affects Cntr-GAN the most, and also the effectiveness of the other augmentations we consider in this paper. 
We conducted extensive experiment on Cntr-GAN with different augmentations 
and report the most representative ones in Figure~\ref{fig:aug_biggan_cntr}.

Overall, we find Cntr-GAN prefers stronger augmentation transforms compared to BCR-GAN.
\textbf{Spatial} augmentations still work better than \textbf{visual} augmentations, which is consistent with our observation that changing the color jittering strength of \emph{simclr} cannot help improve performance.
In Figure~\ref{fig:aug_biggan_cntr}, we present the results of changing the cropping/resizing strength in `simclr', along with the other representative augmentation methods that are helpful to Cntr-GAN.
For most augmentations, CntrGAN reaches the best performance with higher augmentation strength around 0.5.
For CntrGAN, we achieve the best FID of 11.87 applying adjusted \emph{simclr} augmentations with the cropping/resizing strength of 0.3.

\section{Discussion}

Here we provide additional analysis and discussion for several different aspects.
Due to space limit, we summarize our findings below and include visualization of the results in the appendix.

\textbf{Artifacts}.
\citet{ICRGAN} show that imbalanced (only applied to real images) augmentations and regularizations can result in corresponding generation artifacts for GAN models.
Therefore, we present qualitative images sampled randomly for different augmentations and settings of GAN training in the appendix (Section~\ref{sec:appx:artifact}).
For vanilla GAN, augmenting both real and fake images can reduce generation artifacts substantially than only augmenting real images.
With additional contrastive loss and consistency regularization, the generation quality can be improved further.

\textbf{Annealing Augmentation Strength}.
We have extensively experimented with first setting $\lambda_\text{aug}$,
which constrains the augmentation strength, then sampling augmentations randomly.
But how would GANs' performance change if we anneal $\lambda_\text{aug}$ during training?
Our experiments show that annealing the strength of augmentations during training would reduce the effect of the augmentation, 
without changing the relative efficacy of different augmentations.
Augmentations that improve GAN training would alleviate their improvements with annealing;
and vice versa.

\textbf{Composition of Transforms}.
Besides a single augmentation transform, the composition of multiple transforms are also used~\citep{AutoAug,RandAug,AugMix}.
Though the dimension of random composition of transforms is out of this paper's scope,
we experiment with applying both \emph{translation} and \emph{brightness}, as spatial and visual transforms respectively,
to BCR-GAN training.
Preliminary results show that this chained augmentation can achieve the best FID=8.42, while with the single augmentation \emph{translation} the best FID achieved is 8.58,
which suggests this combination is dominant by the more effective \emph{translation}.
We leave it to future work to search for the best strategy of augmentation composition automatically.



\section{Related Work}
Data augmentation has shown to be critical to improve the robustness and generalization of deep learning models, and thus it is becoming an essential component of visual recognition systems ~\citep{AlexNet, ResNet, SIMCLR, FasterRCNN, AugforDetect, DeepLab, GNAE}. More recently, it also becomes one of the most impetus on semi-supervised learning and unsupervised learning~\citep{UDACONSISTENCY, berthelot2019mixmatch, sohn2020fixmatch, xie2019self, berthelot2019remixmatch, SIMCLR}. The augmentation operations also evolve from the basic random cropping and image mirroring to more complicated strategies including geometric distortions (e.g., changes in scale, translation and rotation), color jittering (e.g, perturbations in brightness, contrast and saturation)~\citep{AutoAug,RandAug,ADVAutoAug,AugMix} and  combination of multiple image statistics~\citep{CUTMIX,MIXUP}. 

Nevertheless, these augmentations are still mainly studied in image classification tasks. As for image augmentations in GANs~\citep{GAN}, the progress is very limited: from DCGAN~\citep{DCGAN} to BigGAN~\citep{BIGGAN} and StyleGAN2~\citep{styleganv2}, the mainstream work is only using random cropping and horizontal flipping as the exclusive augmentation strategy. It remains unclear to the research community whether other augmentations can improve quality of generated samples. 
Recently, \citet{MIXUP} stabilized GAN training by mixing both the input and the label for real samples and generated ones. \citet{NOISE} added Gaussian noise to the input images and annealed its strength linearly during the training to achieve better convergence of GAN models. \citet{ArjovskyB17} derived the same idea independently from a theoretical perspective. 
They have shown adding Gaussian noise to both real and fake images can alleviate training instability when the support of data distribution and model distribution do not overlap. \citet{salimans2016improved} further extended the idea by adding Gaussian noise to the output of each layer of the discriminator. 
\citet{Ali2020} found data augmentation improves steerability of GAN models, but they failed to generate realistic samples on CIFAR-10 when jointly optimizing the model and linear walk parameters. 
Besides simply adding augmentation to the data, some recent work~\citep{ChenZRLH19, CRGAN,ICRGAN} further added the regularization on top of augmentations to improve the model performance. For example, Self-Supervised GANs~\citep{ChenZRLH19, LucicTRZBG19} make the discriminator to predict the angle of rotated images and CRGAN~\citep{CRGAN} enforce consistency for different image perturbations.

\section{Conclusion}

In this work, we have conducted a thorough analysis on the performance of different augmentations for improving generation quality of GANs. 
We have empirically shown adding the augmentation to both real images and generated samples is critical for producing realistic samples. 
Moreover, we observe that applying consistency regularization onto augmentations can further boost the performance and it is superior to applying contrastive loss. 
Finally, we achieve state-of-the-art image generation performance by combining constrastive loss and consistency loss. 
We hope our findings can lay a solid foundation and help ease the research in applying augmentations to wider applications of GANs.

\clearpage



\section*{Acknowledgments}
The authors would like to thank 
Marvin Ritter,
Xiaohua Zhai,
Tomer Kaftan,
Jiri Simsa,
Yanhua Sun,
and Ruoxin Sang
for support on questions of codebases;
as well as 
Abhishek Kumar, 
Honglak Lee,
and Pouya Pezeshkpour
for helpful discussions.

\bibliography{main}
\bibliographystyle{plainnat}

\appendix
\clearpage

\section{Notations} \label{sec:appx:notation}

\begin{itemize}

\item $(H, W)$: height and width of an image $\img$

\item $\lambda_\text{aug}$: augmentation strength

\item $\mathcal{U}(\cdot,\cdot)$, $\mathcal{N}(\cdot,\cdot)$, and $\mathcal{B}(\cdot,\cdot)$: Uniform, Gaussian, and Beta distributions respectively. 

\item CR: consistency regularization~\citep{CRGAN}

\item \bcr: balanced consistency regularization~\citep{ICRGAN}

\item \cntr: contrastive loss~\citep{SIMCLR}

\item $\G{z}$: generator output

\item $\D{x}$: discriminator output

\item $\Dh{x}$: hidden representation of discriminator output

\item $\proj{\Dh{x}}$: projection head on top of hiddent representation

\item $\T$: augmentation transforms
\end{itemize}

\section{Augmentations}\label{sec:appx:aug}

\textbf{ZoomIn}: 
We sample $\alpha \sim \mathcal{U}(0, \augstr)$, randomly crop size $((1-\alpha) H, (1-\alpha) W)$ of the image, and resize the cropped image back to $(H,W)$ with bilinear interpolation.

\textbf{ZoomOut}: 
We sample $\alpha \sim \mathcal{U}(0, \augstr)$, 
evenly pad the image to size $((1+2\alpha) H, (1+2\alpha) W)$ with reflection,
randomly crop size $((1+\alpha) H, (1+\alpha) W)$ of the image, 
and resize the cropped image back to $(H,W)$ with bilinear interpolation.

\textbf{TranslationX}:
We sample $\alpha \sim \mathcal{U}(-\augstr, \augstr)$, and shift the image horizontally by $|\alpha|H$ in the direction of $\text{sign}(\alpha)$ with reflection padding.

\textbf{TranslationY}:
We sample $\alpha \sim \mathcal{U}(-\augstr, \augstr)$, and shift the image vertically by $|\alpha|W$ in the direction of $\text{sign}(\alpha)$ with reflection padding.

\textbf{Translation}: 
We sample $\alpha_h, \alpha_w \sim \mathcal{U}(-\augstr, \augstr)$, and shift the image vertically and horizontally by $|\alpha_h|H$ and $|\alpha_w|W$ in the direction of $\text{sign}(\alpha_h)$ and $\text{sign}(\alpha_w)$ with reflection padding.

\textbf{Brightness}: 
We sample $\alpha \sim \mathcal{U}(-\augstr, \augstr)$, add  $\alpha$ to all channels and locations of the image, and clip pixel values to the range $[0, 1]$.

\textbf{Colorness}: 
We sample $\alpha \sim \mathcal{U}(-\augstr, \augstr)$, add  $\alpha$ to one of RGB channels of the image, and clip values to the range $[0, 1]$.

\textbf{InstanceNoise}~\citep{NOISE}:
We add Gaussian noise $\mathcal{N}(0, \augstr)$ to the image.
According to \citet{NOISE}, we also anneal the noise variance from $\augstr$ to 0 during training.

\textbf{CutOut}~\citep{CUTOUT}:
We sample $\alpha \sim \mathcal{U}(0, \augstr)$, and randomly mask out a $(\alpha H, \alpha W)$ region of the image with pixel value of 0.

\textbf{CutMix}~\citep{CUTMIX}:
We sample $\alpha \sim \mathcal{U}(0, \augstr)$, 
pick another random image $\img_1$ in the same batch, 
cut a patch of size $(\alpha H, \alpha W)$ from $\img_1$, 
and paste the patch to the corresponding region in $\img_0$.

\textbf{MixUp}~\citep{MIXUP}:
We first sample $\alpha \sim \mathcal{B}(\augstr, \augstr)$, and set $\alpha = \max (\alpha, 1-\alpha)$. 
Then we pick another random image $\img_1$ in the same batch, 
and use $t(\img_0) = \alpha \img_0 + (1-\alpha) \img_1$ as the augmented image.

\textbf{SimCLR}~\citep{SIMCLR}:
For the default \emph{simclr} augmentation applied to our Cntr-GAN,
we adopt the exact augmentations applied to CIFAR-10 in the opensource code~\footnote{\url{https://github.com/google-research/simclr}} of \citep{SIMCLR}.
The default \emph{simclr} first crops the image with aspect ration in range [3/4, 4/3] and covered area in range [0.08, 1.0].
Then the crops are resized to the original image size, and applied with random horizontal flip.
Finally, color jitters are applied changing the brightness, contrast, saturation, and hue of images.
Please check code opensourced by~\citet{SIMCLR} for more details.

\clearpage
\section{BCR-GAN: GAN with Balanced Consistency Regularization}\label{sec:appx:bcr}
\begin{algorithm}[htb]
\caption{Balanced Consistency Regularized GAN (\bcr-GAN)~\citep{ICRGAN}}
\label{alg:bcr}
\begin{algorithmic}
    \STATE {\bfseries Input:} parameters of generator $\theta_G$ and discriminator $\theta_D$, consistency regularization coefficient $\lambda_\bcr$, augmentation transforms $\T$, assuming the discriminator updates only once per generator iteration.
    \FOR{number of training iterations}
    \STATE Sample batch $x \sim p_{\text{real}}(x)$, $z \sim p(z)$
    \STATE Real images $\img_\real = x$, fake images $\img_\fake = \G{z}$    
    \STATE $L_D \leftarrow D(\img_\fake) - D(\img_\real)$
    \STATE Sample augmentation transforms $t_1 \sim \T, t_2 \sim \T$
    \STATE Augment both real and fake images $t_1(\img_\real), t_2(\img_\fake)$    
    \STATE $L_\bcr \leftarrow \lVert D(\img_\real) - D(t_1(\img_\real)) \rVert ^2 + \lVert D(\img_\fake) - D(t_2(\img_\fake)) \rVert ^2$
    \STATE $\theta_D \leftarrow \text{AdamOptimizer}(L_D + \lambda_\bcr L_\bcr)$
    \STATE $L_G \leftarrow -D(G(z))$
    \STATE $\theta_G \leftarrow \text{AdamOptimizer}(L_G)$
    \ENDFOR
\end{algorithmic}
\end{algorithm}

\section{Cntr-GAN: GAN with Contrastive Loss}\label{sec:appx:cntr}

We first elaborate on contrastive loss as defined in \citet{SIMCLR}.
Given a minibatch representations of $N$ examples, and another minibatch representations of corresponding augmented examples, we concatenate them into a batch of $2N$ examples.
After concatenation, a positive pair $(i,j)$ should have $|i-j|=N$, 
while we treat the other $2(N-1)$ samples within the batch as negative examples.
Then the loss function for a positive pair of examples $(i, j)$ is defined as:
\begin{align} 
\ell_{i, j} &= -\log \frac{\exp(\text{sim}(h_i, h_j)/\tau)}
{\sum_{k=1}^{2N} \mathbbm{1}_{[k \neq i]} \exp(\text{sim}(h_i, h_k)/\tau)}  \nonumber
\end{align}
where $\text{sim}(u, v) = u^T v / \Vert u \Vert \Vert v \Vert$ denotes the cosine similarity between two vectors, $\mathbbm{1}_{[k \neq i]} \in \{0,1\}$ is an indicator evaluating to 1 iff $k \neq i$, and $\tau$ denotes a temperature hyper-parameter.
The final contrastive loss (\textbf{CntrLoss}) is computed across all positive pairs, both $(i,j)$ and $(j,i)$ in the concatenated batch of size $2N$.

Then we propose Cntr-GAN, in which we apply contrastive loss to GANs with both real and fake images augmented during training. 
Algorithm~\ref{alg:cntr} details how we augment images and regularize GAN training with CntrLoss.

\begin{algorithm}[htb]
\caption{Contrastive Loss regularized GAN (\cntr-GAN)}
\label{alg:cntr}
\begin{algorithmic}
    \STATE {\bfseries Input:} parameters of generator $\theta_G$ and discriminator $\theta_D$, $\Dh$ returns the last hidden representation from the discriminator, $\proj$ is the projection head that maps representations to the space where contrastive loss is applied, contrastive loss coefficient $\lambda_\cntr$, augmentation transforms $\T$, assuming the discriminator updates only once per generator iteration.
    \FOR{number of training iterations}
    \STATE Sample batch $x \sim p_{\text{real}}(x)$, $z \sim p(z)$
    \STATE Real images $\img_\real = x$, fake images $\img_\fake = \G{z}$    
    \STATE $L_D \leftarrow D(\img_\fake) - D(\img_\real)$
    \STATE Sample augmentation transforms $t_1 \sim \T, t_2 \sim \T, t_3 \sim \T, t_4 \sim \T$
    \STATE Augment both real and fake images $t_1(\img_\real), t_2(\img_\real), t_3(\img_\fake), t_4(\img_\fake)$
    \STATE Obtain projected representations $h_{\real1} = \proj{\Dh{t_1(\img_\real)}}, h_{\real2} = \proj{\Dh{t_2(\img_\real)}}$
    \STATE $h_{\fake1} = \proj{\Dh{t_3(\img_\fake)}}, h_{\fake2} = \proj{\Dh{t_4(\img_\fake)}}$
    \STATE $L_\cntr \leftarrow \text{CntrLoss}(h_{\real1}, h_{\real2}) + \text{CntrLoss}(h_{\fake1}, h_{\fake2})$
    \STATE $\theta_D \leftarrow \text{AdamOptimizer}(L_D + \lambda_\cntr L_\cntr)$
    \STATE $L_G \leftarrow -D(G(z))$
    \STATE $\theta_G \leftarrow \text{AdamOptimizer}(L_G)$
    \ENDFOR
\end{algorithmic}
\end{algorithm}

\clearpage

\section{Generation Artifacts} \label{sec:appx:artifact}

\begin{figure}[ht]
\begin{subfigure}{0.49\linewidth}
\frame{\includegraphics[width=\linewidth]{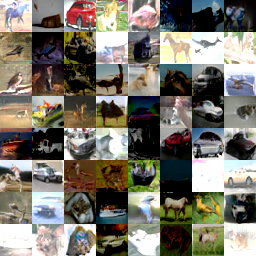}}
\caption*{BigGAN w/ only $\img_\real$ augmented, FID=29.23}\end{subfigure} \hfill
\begin{subfigure}{0.49\linewidth}
\frame{\includegraphics[width=\linewidth]{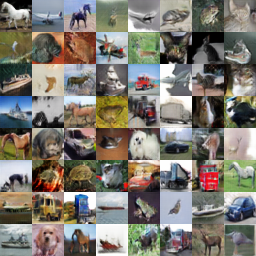}}
\caption*{BigGAN w/ $\img_\real$ \& $\img_\fake$ augmented, FID=14.02}\end{subfigure}\\
\begin{subfigure}{0.49\linewidth}
\frame{\includegraphics[width=\linewidth]{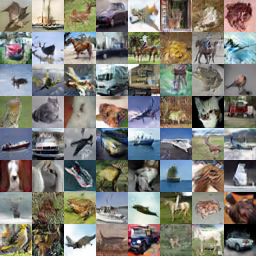}}
\caption*{Cntr-BigGAN w/ $\img_\real$ \& $\img_\fake$ augmented, FID=13.75}\end{subfigure} \hfill
\begin{subfigure}{0.49\linewidth}
\frame{\includegraphics[width=\linewidth]{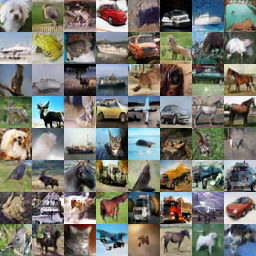}}
\caption*{BCR-BigGAN w/ $\img_\real$ \& $\img_\fake$ augmented, FID=13.19}\end{subfigure}
\caption{Random qualitative examples showing brightness artifacts with $\lambda_\text{aug}=0.3$.
For vanilla GAN, augmenting both real and fake images can reduce generation artifacts substantially than only augmenting real images.
With additional contrastive loss and consistency regularization, the generation quality can be improved further.
}
\label{fig:artifacts:brightness}
\end{figure}

\begin{figure}[ht]
\begin{subfigure}{0.49\linewidth}
\frame{\includegraphics[width=\linewidth]{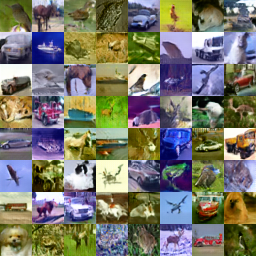}}
\caption*{BigGAN w/ only $\img_\real$ augmented, FID=33.26}\end{subfigure} \hfill
\begin{subfigure}{0.49\linewidth}
\frame{\includegraphics[width=\linewidth]{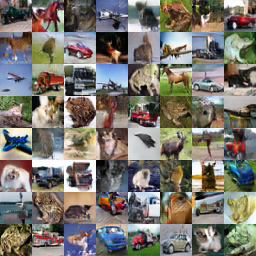}}
\caption*{BigGAN w/ $\img_\real$ \& $\img_\fake$ augmented, FID=16.52}\end{subfigure}\\ 
\begin{subfigure}{0.49\linewidth}
\frame{\includegraphics[width=\linewidth]{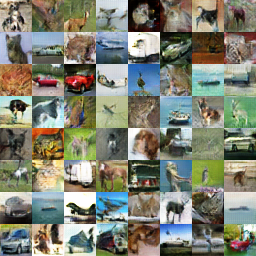}}
\caption*{Cntr-BigGAN w/ $\img_\real$ \& $\img_\fake$ augmented, FID=15.57}\end{subfigure} \hfill
\begin{subfigure}{0.49\linewidth}
\frame{\includegraphics[width=\linewidth]{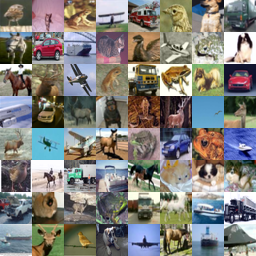}}
\caption*{BCR-BigGAN w/ $\img_\real$ \& $\img_\fake$ augmented, FID=14.41}\end{subfigure}
\caption{Random qualitative examples showing blueness artifacts with $\lambda_\text{aug}=0.3$.
For vanilla GAN, augmenting both real and fake images can reduce generation artifacts substantially than only augmenting real images.
With additional contrastive loss and consistency regularization, the generation quality can be improved further.
}
\label{fig:artifacts:blueness}
\end{figure}

\begin{figure}[ht]
\begin{subfigure}{0.49\linewidth}
\frame{\includegraphics[width=\linewidth]{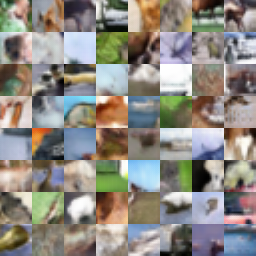}}
\caption*{BigGAN w/ only $\img_\real$ augmented, FID=50.98}\end{subfigure} \hfill
\begin{subfigure}{0.49\linewidth}
\frame{\includegraphics[width=\linewidth]{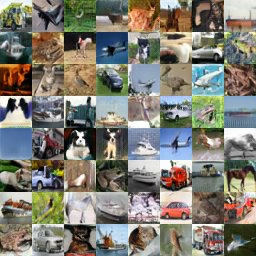}}
\caption*{BigGAN w/ $\img_\real$ \& $\img_\fake$ augmented, FID=14.65}\end{subfigure}\\ 
\begin{subfigure}{0.49\linewidth}
\frame{\includegraphics[width=\linewidth]{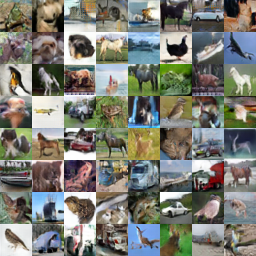}}
\caption*{Cntr-BigGAN w/ $\img_\real$ \& $\img_\fake$ augmented, FID=13.08}\end{subfigure} \hfill
\begin{subfigure}{0.49\linewidth}
\frame{\includegraphics[width=\linewidth]{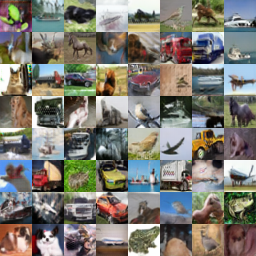}}
\caption*{BCR-BigGAN w/ $\img_\real$ \& $\img_\fake$ augmented, FID=11.66}\end{subfigure}
\caption{Random qualitative examples showing zoomin artifacts with $\lambda_\text{aug}=0.3$.
For vanilla GAN, augmenting both real and fake images can reduce generation artifacts substantially than only augmenting real images.
With additional contrastive loss and consistency regularization, the generation quality can be improved further.
}
\label{fig:artifacts:zoomin}
\end{figure}

\begin{figure}[ht]
\begin{subfigure}{0.49\linewidth}
\frame{\includegraphics[width=\linewidth]{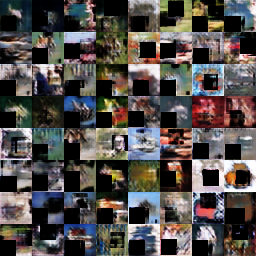}}
\caption*{BigGAN w/ only $\img_\real$ augmented, FID=47.76}\end{subfigure} \hfill
\begin{subfigure}{0.49\linewidth}
\frame{\includegraphics[width=\linewidth]{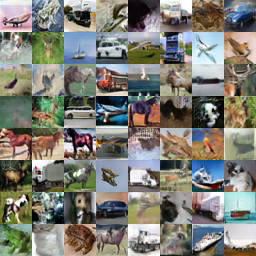}}
\caption*{BigGAN w/ $\img_\real$ \& $\img_\fake$ augmented, FID=14.30}\end{subfigure}\\ 
\begin{subfigure}{0.49\linewidth}
\frame{\includegraphics[width=\linewidth]{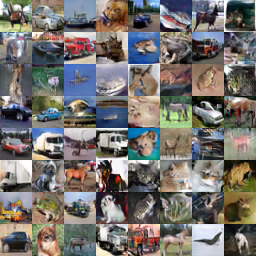}}
\caption*{Cntr-BigGAN w/ $\img_\real$ \& $\img_\fake$ augmented, FID=14.12}\end{subfigure} \hfill
\begin{subfigure}{0.49\linewidth}
\frame{\includegraphics[width=\linewidth]{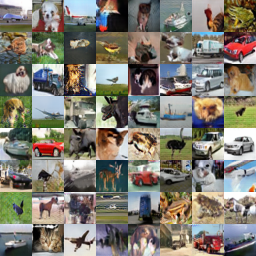}}
\caption*{BCR-BigGAN w/ $\img_\real$ \& $\img_\fake$ augmented, FID=12.63}\end{subfigure}
\caption{Random qualitative examples showing cutout artifacts with $\lambda_\text{aug}=0.3$.
For vanilla GAN, augmenting both real and fake images can reduce generation artifacts substantially than only augmenting real images.
With additional contrastive loss and consistency regularization, the generation quality can be improved further.
}
\label{fig:artifacts:cutout}
\end{figure}

\clearpage
\section{Additional Results}\label{sec:appx:res}

\subsection{BigGAN with Only Real Images Augmented} \label{sec:appx:vanilla}

As extra results for Section~\ref{sec:vanilla_r}, the results of BigGAN with only real images augmented is consistent with Figure~\ref{fig:aug_only_real_vanilla}. 
It further shows only augmenting real images is not helpful with vanilla GAN.

\begin{figure}[hb] 
  \centering
    \includegraphics[width=1\linewidth]{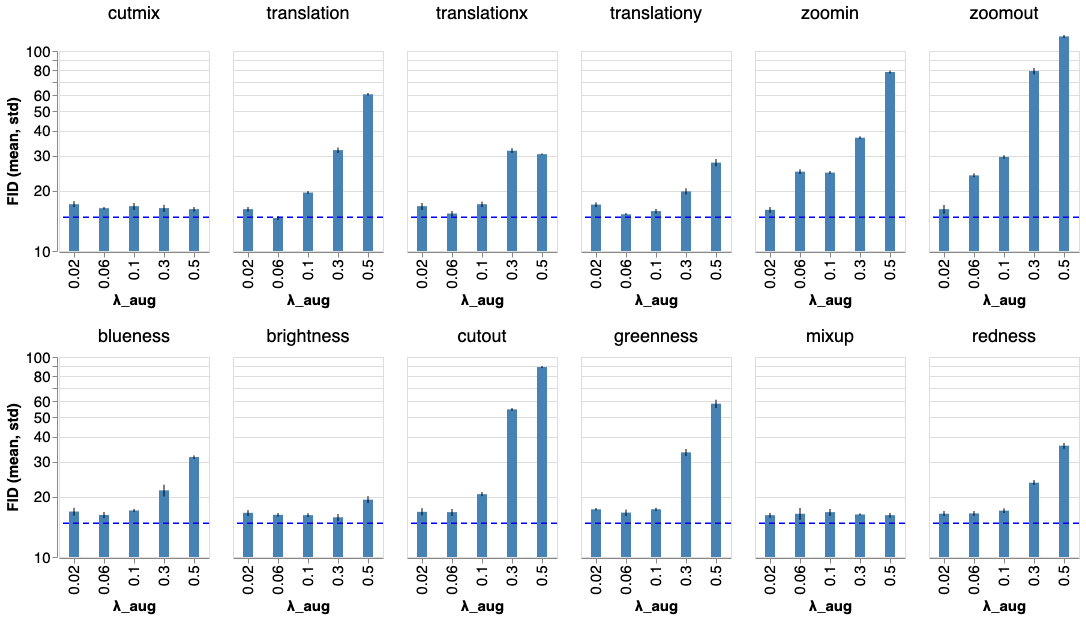}
  \caption{FID comparisons of BigGAN trained on augmented real images only.}
\label{fig:aug_only_real_vanilla_biggan}
\end{figure}

\subsection{Interaction between CntrLoss and BCR} \label{sec:appx:cntr_bcr}

In Section~\ref{sec:cntr},
we experiment with applying both CntrLoss and BCR to regularizing BigGAN.
We achieve new state-of-the-art $\text{FID}=8.30$ with the strength of CntrLoss $\lambda_\cntr=0.1$ and the strength of BCR $\lambda_\bcr=5$.
While BCR enforces the consistency loss directly on the discriminator logits,
with Cntr together, it further helps to learn better representations which can be reflected in generation performance eventually.

\begin{figure}[hb]
  \centering
    \includegraphics[width=0.6\linewidth]{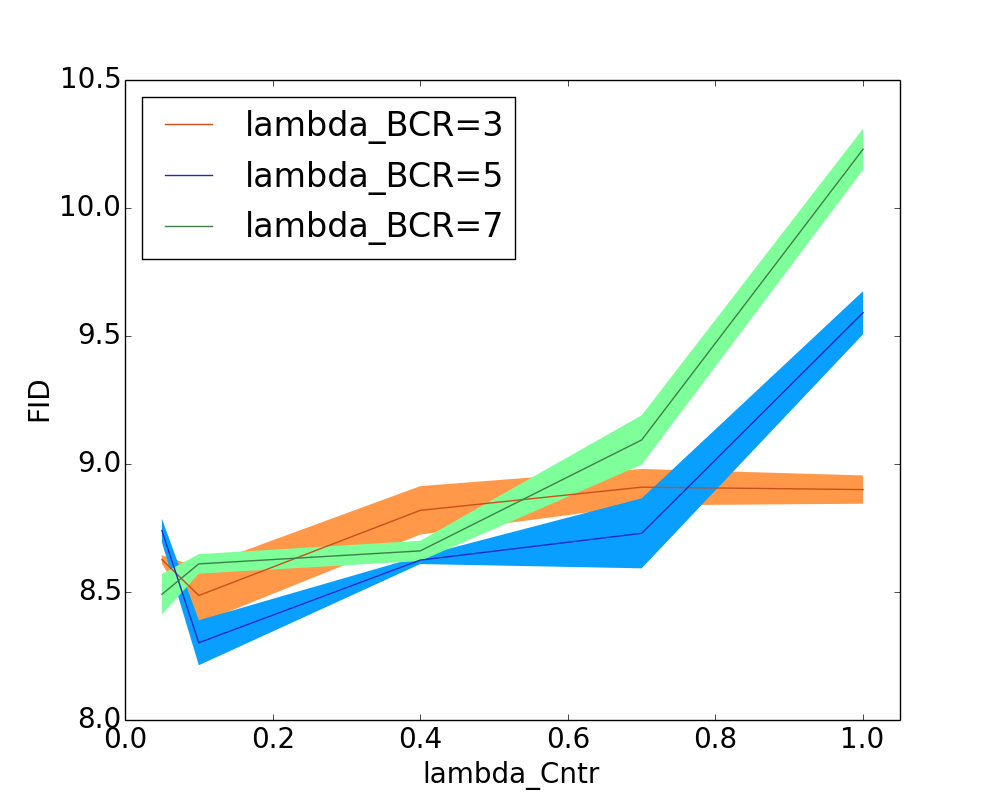}
  \caption{BigGAN on CIFAR-10 regularized with both Cntr and BCR. We achieve new state-of-the-art $\text{FID}=8.30$ with $\lambda_\cntr=0.1, \lambda_\bcr=5$.
  }
 \label{fig:cntr_bcr}
\end{figure}

\clearpage
\subsection{Annealing Augmentation Strength during Training}
Our experiments show that annealing the strength of augmentations during training would reduce the effect of the augmentation, 
without changing the relative efficacy of different augmentations.
Augmentations that improve GAN training would alleviate their improvements with annealing;
and vice versa.

\begin{figure}[hb] 
  \centering
    \includegraphics[width=1\linewidth]{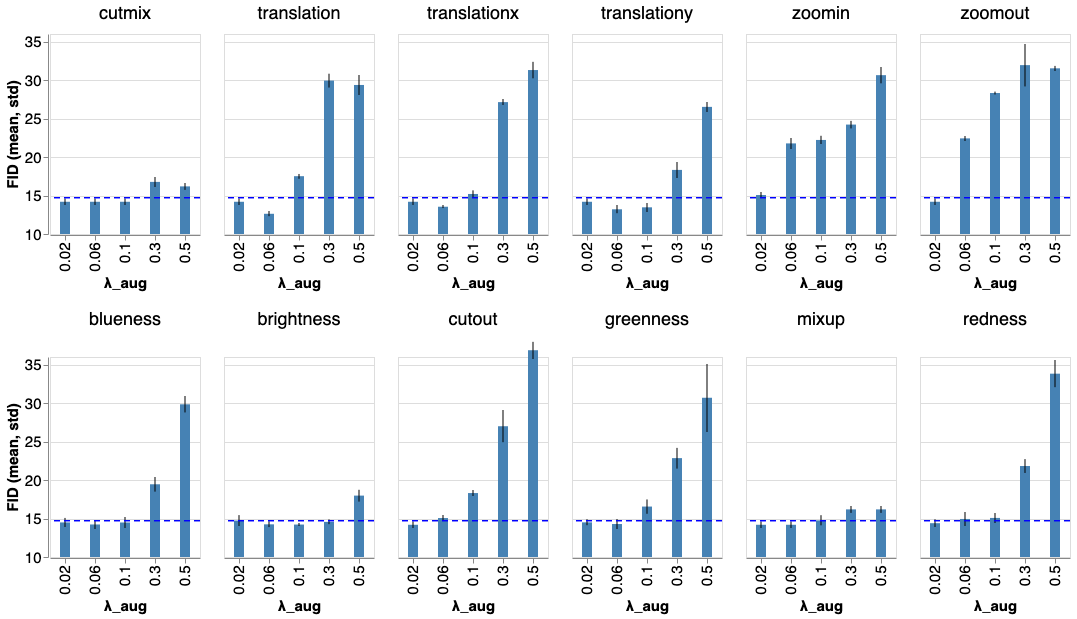}
  \caption{Annealing Augmentation Strength during Training.}
\label{fig:anneal}
\end{figure}

\subsection{Exploration on Chain of Augmentations}
We experiment with applying both \emph{translation} and \emph{brightness}, as spatial and visual transforms respectively,
to BCR-GAN training.
Preliminary results show that this chained augmentation can achieve the best FID=8.42, while with the single augmentation \emph{translation} the best FID achieved is 8.58.
This suggests the combination of \emph{translation} and \emph{brightness} is dominant by the more effective \emph{translation}.
We leave it to future work to search for the best strategy of augmentation composition automatically.

\begin{figure}[ht]
  \centering
    \includegraphics[width=0.6\linewidth]{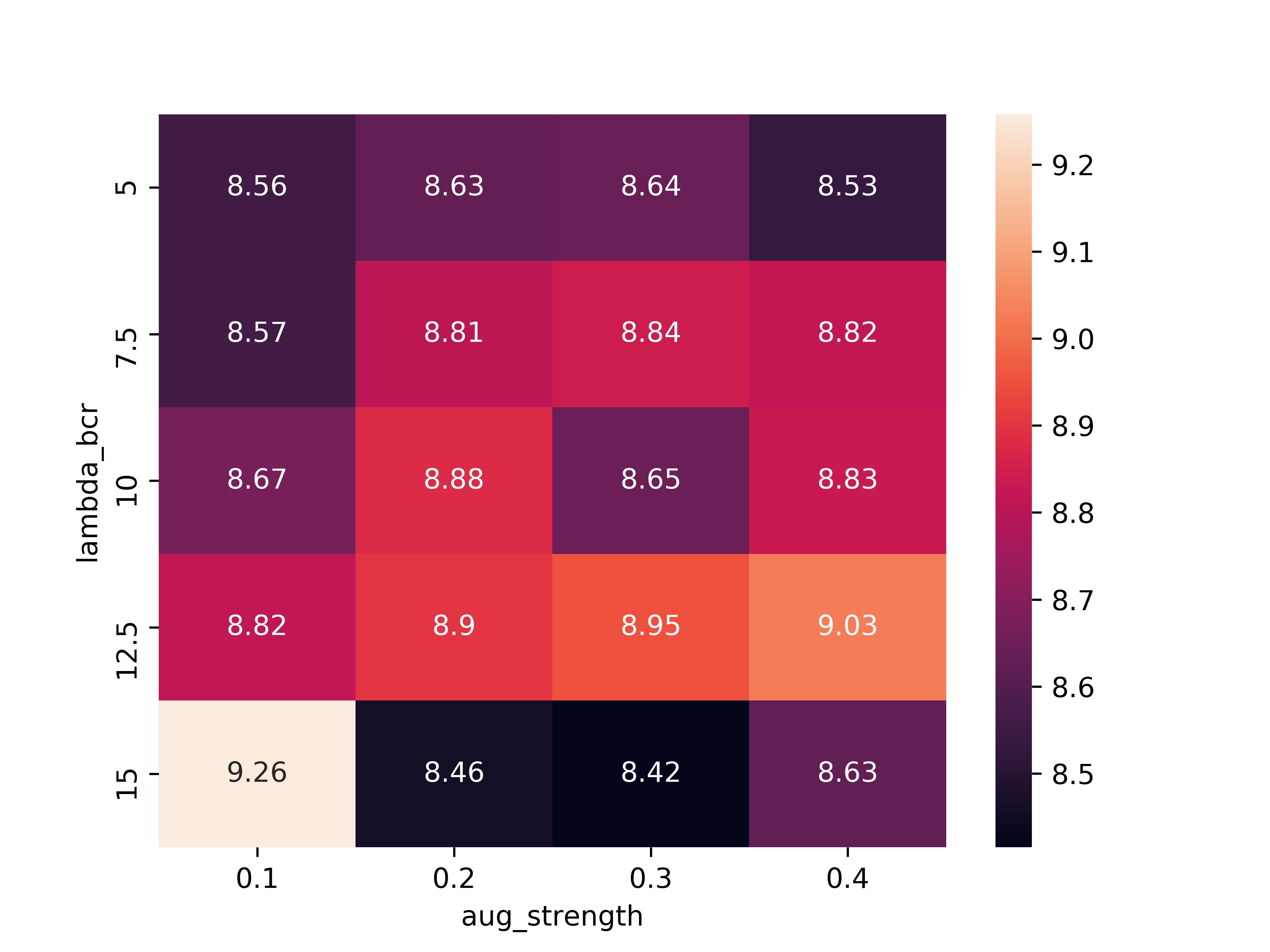}
  \caption{BCR with \emph{translation} + \emph{brightness}
  }
 \label{fig:bcr_translation_brightness}
\end{figure}

\clearpage
\section{Model Details}

\paragraph{Unconditional SNDCGAN}
The SNDCGAN architecture is shown below. Please refer~\citet{kurach2019large} for more details.

\begin{figure}[ht]
  \centering
    \includegraphics[width=\linewidth]{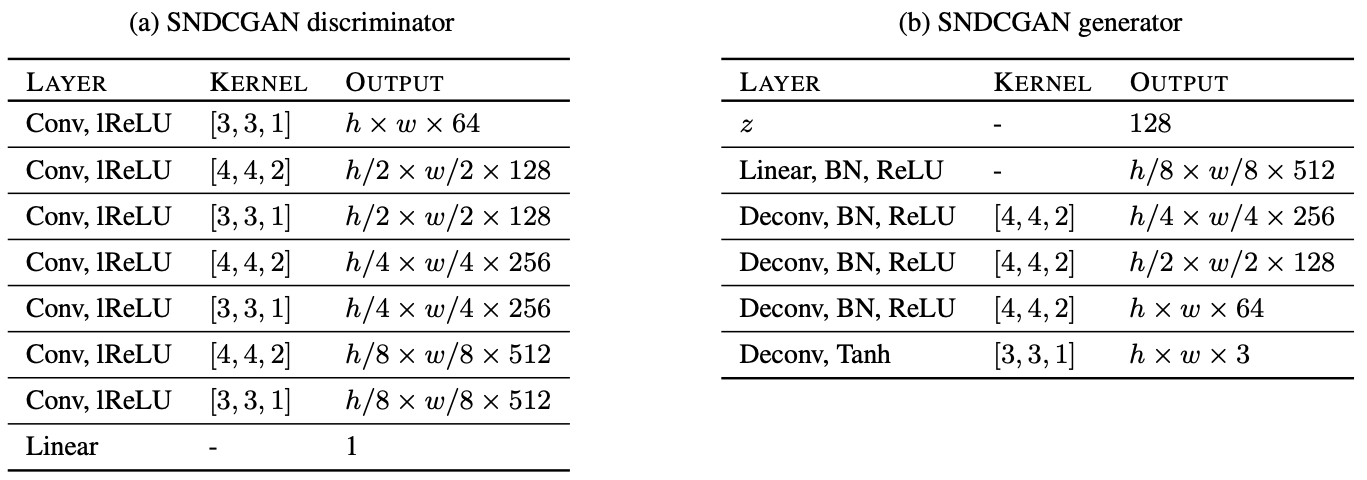}
 \label{fig:sndcgan_arch}
\end{figure}

\paragraph{Conditional BigGAN}
The BigGAN architecture is shown below. Please refer~\citet{BIGGAN} for more details.

\begin{figure}[ht]
  \centering
    \includegraphics[width=0.8\linewidth]{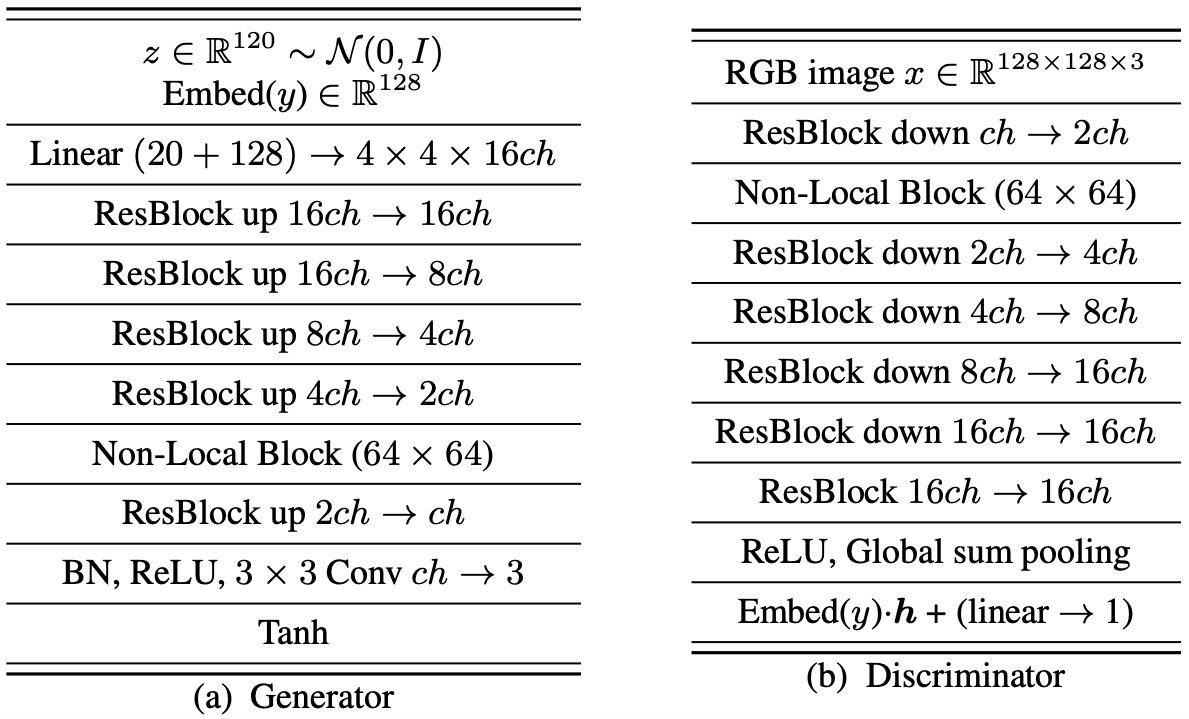}
 \label{fig:biggan_arch}
\end{figure}

\end{document}